\documentclass{article} % For LaTeX2e
\usepackage{iclr2026_conference,times}
\usepackage{graphicx}
\usepackage{hyperref}
\usepackage{amsmath,amsfonts}
\usepackage{array}
\usepackage{textcomp}
\usepackage{stfloats}
\usepackage{url}
\usepackage{verbatim}
\usepackage{amssymb}
\usepackage{multirow}
\usepackage{booktabs}
\usepackage{subcaption}
\usepackage{colortbl}
\usepackage{bbding}
\usepackage{wrapfig}
\usepackage{balance} 
\usepackage[table]{xcolor}
\usepackage[most]{tcolorbox} % 核心宏包
\usepackage{xcolor}          % 用于定义自定义颜色
\definecolor{oursburgundy}{HTML}{DCEEFF}
\definecolor{warmback}{RGB}{254, 252, 240}   % 极浅的米色
\definecolor{warmframe}{RGB}{225, 215, 190}  % 沙色边框
\definecolor{warmtitle}{RGB}{140, 60, 40}    % 深赤褐色标题

\newtcolorbox{querybox}{
    colback=warmback,
    colframe=warmframe,
    coltext=black,
    boxrule=0.6pt,
    arc=0pt, 
    left=10pt,              % 左缩进
    right=5pt,             % 右缩进
    top=8pt,                % 上边距
    bottom=8pt,             % 下边距
    enhanced,               % 启用增强模式
    breakable               % 允许跨页（如果内容很多的话）
}

\title{Causal-EVC: Breaking Emotional Spurious Causality via Spatiotemporal Grounding and Counterfactual Intervention}

\author{Cheng Ye$^1$, Weidong Chen$^1$, Peipei Song$^1$, Zhendong Mao$^1$ \\
$^1$University of Science and Technology of China, Hefei \\
\texttt{chenweidong@ustc.edu.cn}
}
\iclrfinalcopy % Uncomment for camera-ready version, but NOT for submission.
\begin{document}

\maketitle

\begin{abstract}
Emotional Video Captioning aims to generate factually accurate and emotionally empathetic descriptions. While recent methods have recognized the importance of visual causes to guide emotion perception and caption generation, they fundamentally rely on simple attention matching, which inevitably suffers from {causal redundancy and spurious correlations} in co-occurrence bias (\emph{e.g.}, misclassifying ``sadness'' as ``joy'' on a sunny beach), leading to severe shortcut learning from confusing backgrounds. Furthermore, existing evaluations fail to verify whether models have genuinely mastered causal reasoning or merely exploited background confounders. To address these limitations, we first construct {EVC-CauseGround}, a comprehensive benchmark with dense spatio-temporal causal annotations. Crucially, it introduces a carefully selected {Causal-Faithfulness Subset} to explicitly quantify genuine emotion-cause attribution. Second, we propose {Causal-EVC}, an emotion-grounding captioning framework, which introduces a Motion-guided Causal Spatiotemporal Localization module to precisely decouple causal triggers from background confounders. Besides, we introduce an Interpretable Sparse Emotion Routing module. By synthesizing counterfactual representations and formulating a novel counterfactual contrastive objective, we enforce the model to anchor its emotion predictions strictly on authentic causal triggers instead of confusing background. Extensive experiments show that Causal-EVC not only achieves the best performance on semantic metrics but also exhibits significant advantages in the causal-faithfulness subset, which demonstrates that our model could mine emotional cues from genuine visual causes and mitigate co-occurrence bias for interpretable multimodal emotion understanding.\footnote{Code will be released in the final version of the paper.}
\end{abstract}

% With the rapid evolution of self-media platforms, users are no longer satisfied with mere information transmission, they are increasingly inclined to express subjective emotions through multimodal contents. This burgeoning demand for emotional expression has catalyzed the community’s enthusiasm for exploring automated visual emotion analysis, \emph{i.e.,} facial emotion recognition~\citep{fard2025affectnet+,lan2025expllm,guo2024benchmarking,chen2026subjective}, video emotion classification ~\citep{wang2025mdkat,li2025feature,qi2024versatile,ye2025multi,wang2025combatting}, and emotional video captioning (EVC) ~\citep{liu2025enriched, yu2024prompting,yu2023comprehensive, deng2021syntax}. Among them, Emotional Video Captioning (EVC) task, which aims to generate factually accurate and emotionally rich descriptions for video, stands out as a fundamental yet highly challenging endeavor due to the complex intersection of vision, emotion, and language studies. Firstly, EVC necessitates the effective alignment and fusion of content across multiple modalities (e.g., vision and text). Besides, it requires to not only comprehend factual content but also accurately perceive implicit emotional cues within the video and seamlessly integrate them into caption generation. EVC offers significant potential for various real-world applications, \emph{i.e.,} empathetic conversational agent, video comment generation, and automated film production.
\section{Introduction}
{Emotional Video Captioning (EVC)} aims to generate factually accurate and emotionally empathetic descriptions for videos~\citep{el2026hume,deng2026pixels,zhang2026mme}. Unlike standard video captioning~\citep{ye2025improving,li2026if,xu2026carebench,chen2026subjective}, EVC represents a fundamentally more complex endeavor. It requires models to not only cross-modally align factual visual content with natural language, but also accurately perceive implicit affective cues and seamlessly synthesize them into human-like narratives.

\begin{figure*}[t]        
\center{\includegraphics[width=1\linewidth] {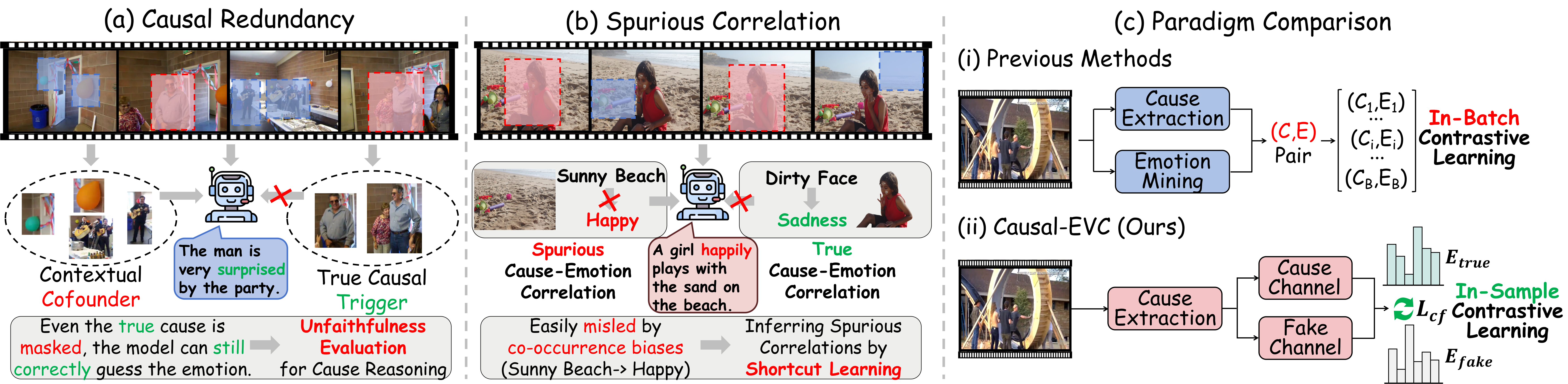}}
\caption{Motivation. Traditional methods suffer from (a) Causal Redundancy and (b) Spurious Correlations, where models exploit shortcut learning and emotional hallucinations due to background confounders. (c) Paradigm Comparison between Causal-EVC and previous methods demonstrate the superiority for genuine emotion-cause reasoning of Causal-EVC.}
\label{fig1}
\vspace{-20pt}
\end{figure*}

Despite recent progress, mainstream EVC methods~\citep{wang2021emotion,song2022contextual,xu2024cross} predominantly map global video representations to predefined emotion spaces via attention patterns. However, this paradigm overlooks the intrinsic nature of visual emotions, which are typically triggered by fine-grained motivational causes rather than global backgrounds. While pioneering efforts like MM-ECPE~\citep{ye2025multi} attempted to explicitly extract emotion-cause pairs, we argue that current cause attribution methods still suffer from two fundamental limitations:
\textbf{1) Causal Redundancy under Consistent Backgrounds:} As shown in Fig. 1(a), when the contextual cofounder semantically aligns with the true emotion, the background inherently leaks the emotion and generates the causal redundancy. Because MM-ECPE merely maximizes positive mutual information, models effortlessly exploit these environmental confounders to achieve high prediction accuracy while ignoring the authentic causal triggers, failing to penalize this shortcut learning and losing the genuine emotional reasoning capabilities.
\textbf{2) Spurious Correlations under Inconsistent Backgrounds:} Conversely, when the background conflicts with the true foreground trigger (\emph{e.g.}, a child crying on a sunny beach), traditional patterns are heavily hacked by dataset-level co-occurrence bias (\emph{e.g.}, ``Sunny Beach $\rightarrow$ Joy''). Lacking causal intervention mechanisms, the model is easily misled by the dominant confounder and produces severe emotional hallucinations.
Meanwhile, existing benchmarks only evaluate the quality of generated captions and the accuracy of the predicted emotions, which completely fails to quantify the true causal reasoning ability and prevent making lucky statistical guesses based on environmental leakage.

  \begin{wrapfigure}{r}{0.52\linewidth} 
  \vspace{-10pt}
     \centering
     \includegraphics[width=0.95\linewidth]{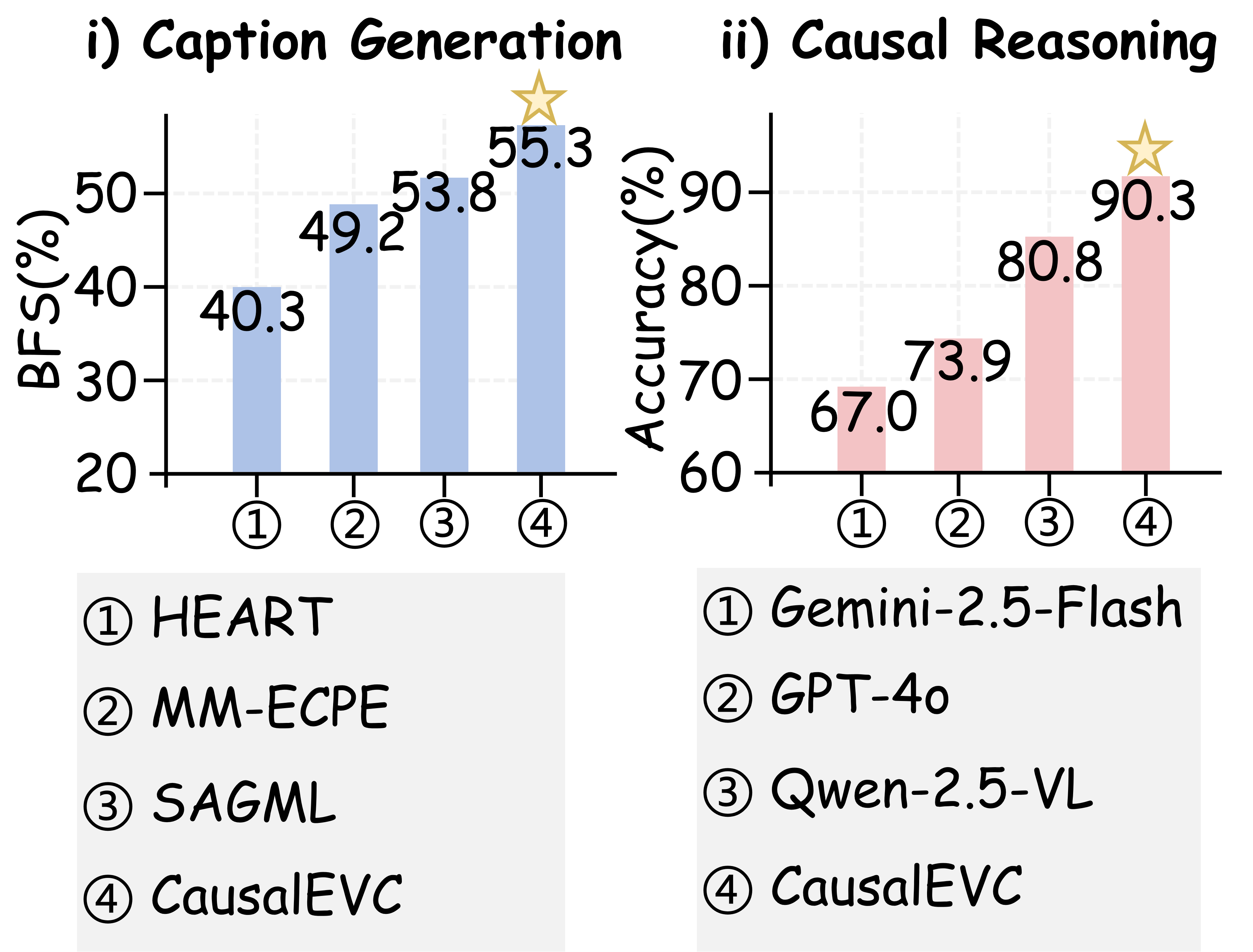} 
     \caption*{Figure 2: Performance Comparison on the caption generation and causal reasoning.}
     \label{fig11}
      \vspace{-10pt}
 \end{wrapfigure}

To fundamentally resolve these dilemmas, we tackle the EVC challenge from both benchmark and model perspectives:
1) We construct \textbf{EVC-CauseGround}, the first benchmark dedicated to fine-grained emotion causality. We achieve this via two major enhancements: We first introduce dense spatio-temporal causal annotations, defining precise temporal spans and bounding-box tubes for visual causes. Crucially, we collect a pure {Causal-Faithfulness Subset}, where background leakage is strictly eliminated. Based on it, we design a novel metric $\text{ST-IoU}$ to rigorously quantify genuine emotion attribution.
2) We propose \textbf{Causal-EVC}, a counterfactual causal grounding framework. Technically, we design a Motion-Guided Causal Spatiotemporal Localization module to guide spatial and temporal attention, generating an accurate causal mask. Guided by it, we physically sever the causal path by synthesizing counterfactual background representations and propose an Interpretable Sparse Emotion Routing module to deduce class-specific emotion probabilities. By enforcing a Counterfactual Contrastive Objective, we explicitly penalize background exploitation, ensuring emotions are reasoned strictly from foreground triggers. As shown in Fig. 2, our model achieves consistent improvements in both caption generation and causal reasoning. In short, our main contributions are summarized as follows:

$\bullet$ \textbf{Benchmark}: We construct {EVC-CauseGround}, a pioneering benchmark with dense spatiotemporal annotations to evaluate the causal attribution capabilities. Besides, we design a convincing metric $\text{ST-IoU}$ to explicitly quantify the genuine emotion-cause reasoning ability.

$\bullet$ \textbf{Framework}: We propose {Causal-EVC}, a novel framework with motion-guided cause localization and counterfactual emotion deduction. Besides, we design several objective functions to constrain cause extraction and emotion mining, which breaks spurious correlations and compels the model to reason emotions from true triggers.

$\bullet$ \textbf{Performance}: Extensive experiments demonstrate the superiority of Causal-EVC especially in cause grounding, \emph{i.e.,} +17.3\%/11.8\% improvements on ${\rm mIoU}$ and emotion accuracy than zero-shot Qwen-2.5-VL-7B, which validates that our approach learns genuine emotional causal reasoning capability for better emotional descriptions.

\begin{figure*}[t]        
\center{\includegraphics[width=1\linewidth] {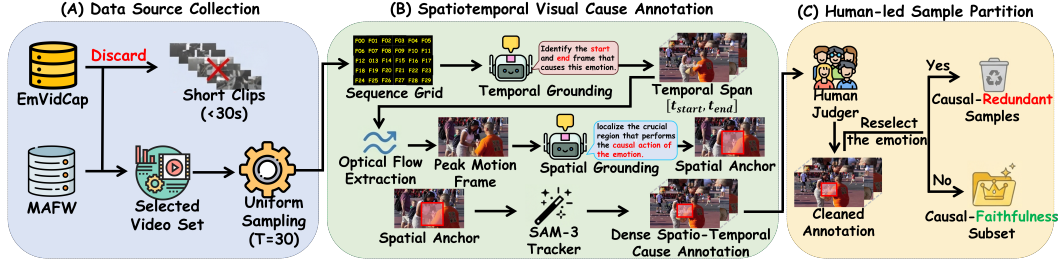}}
\caption*{Figure 3: The construction pipeline of the EVC-CauseGround benchmark, which is consist of (A) Data Source Collection, (B) Spatiotemporal Visual Cause Annotation and (C) Human-led Sample Check and Partition, respectively.}
\label{figbench}
\end{figure*}

\section{EVC-CauseGround: A Visual-Cause Grounding Emotional Descriptive Benchmark}
In this paper, we construct {EVC-CauseGround}, a novel benchmark that extends conventional emotion descriptions with a fine-grained emotional cause annotations. Unlike traditional datasets that provide only holistic labels or coarse utterance-level annotations~\citep{wang2022multimodal,ma2024extraction,zhang2020does}, \textbf{EVC-CauseGround is the first benchmark to unify fine-grained spatio-temporal causal tubes and a rigorously curated counterfactual subset}\footnote{Details and the Pipeline figure of EVC-CauseGround are shown in the Appendix.}, establishing a comprehensive benchmark for authentic emotional cause reasoning.

\textbf{Data Source Collection.}
We meticulously curate diverse, in-the-wild videos from {EmVidCap}\citep{wang2021emotion} and {MAFW}\citep{liu2022mafw} datasets. Specifically, we discard videos shorter than 5 seconds, as too-short durations make fine-grained causal localization trivial. We finally collect 1,523 videos from EmVidCap and 477 videos from MAFW (averaging $\sim$100s), covering unconstrained scenarios from daily interactions to natural disasters. They are split into 1,500 training and 500 evaluation samples. All selected videos are natively equipped with ground-truth emotion labels and captions, serving as reliable anchors for our subsequent causal annotation.

\textbf{Spatiotemporal Visual Cause Annotation.}
To avoid cost-prohibitive manual frame-by-frame drawing, we propose an efficient MLLM-led, tracker-assisted pipeline. We first uniformly sample $T=30$ frames from each video and concatenate them into a $6\times 5$ grid image with sequence identifiers (e.g., $F00$ to $F29$). We prompt an advanced MLLM (e.g., GPT-5.6~\citep{singh2025openai}) with this grid and the ground-truth emotion to deduce the causal temporal window $[t_{start},t_{end}]$. To prevent lazy behaviors of MLLMs that include all frames, we constraint that the causal span must not exceed 30\% of the total frames.
For spatial annotation, we first compute inter-frame optical flow within $[t_{start},t_{end}]$ to identify the \textit{peak motion frame}, which is the climax of the causal action. By feeding only this high-resolution frame into the MLLM, we obtain a highly confident static bounding box, devoid of multi-frame interference. Utilizing the generated anchor box as an initial prompt, we leverage the zero-shot tracker SAM-3~\citep{carion2026sam} to bidirectionally track the causal object across the temporal span, yielding continuous and dense spatio-temporal causal tubes.

\textbf{Human-led Sample Check and Partition.}
While the automated pipeline accelerates annotation, human verification is essential to eliminate hallucinations. Firstly, human annotators are required to check for redundant or missed frames at the boundaries of the temporal spans and spatial regions. Besides, since MLLMs fail to address the \textit{Contextual Leakage} issue, we synthesize a \textit{counterfactual video} for each sample by blacking out the verified causal tube. Then, annotators, blinded to the original labels, are asked to guess the emotion based solely on the masked background.
If annotators can still correctly predict the emotion, the sample is flagged as {Causal-Redundant} (i.e., background confounders are too dominant). Conversely, the video is flagged as {Causal-Faithfulness}. Finally, we collect a \textbf{Causal-Faithfulness Subset} containing 176 videos, which empowers us to evaluate whether a model performs genuine causal reasoning or merely exploits contextual cofounders.

\begin{figure*}[t]
\center{\includegraphics[width=0.98\linewidth] {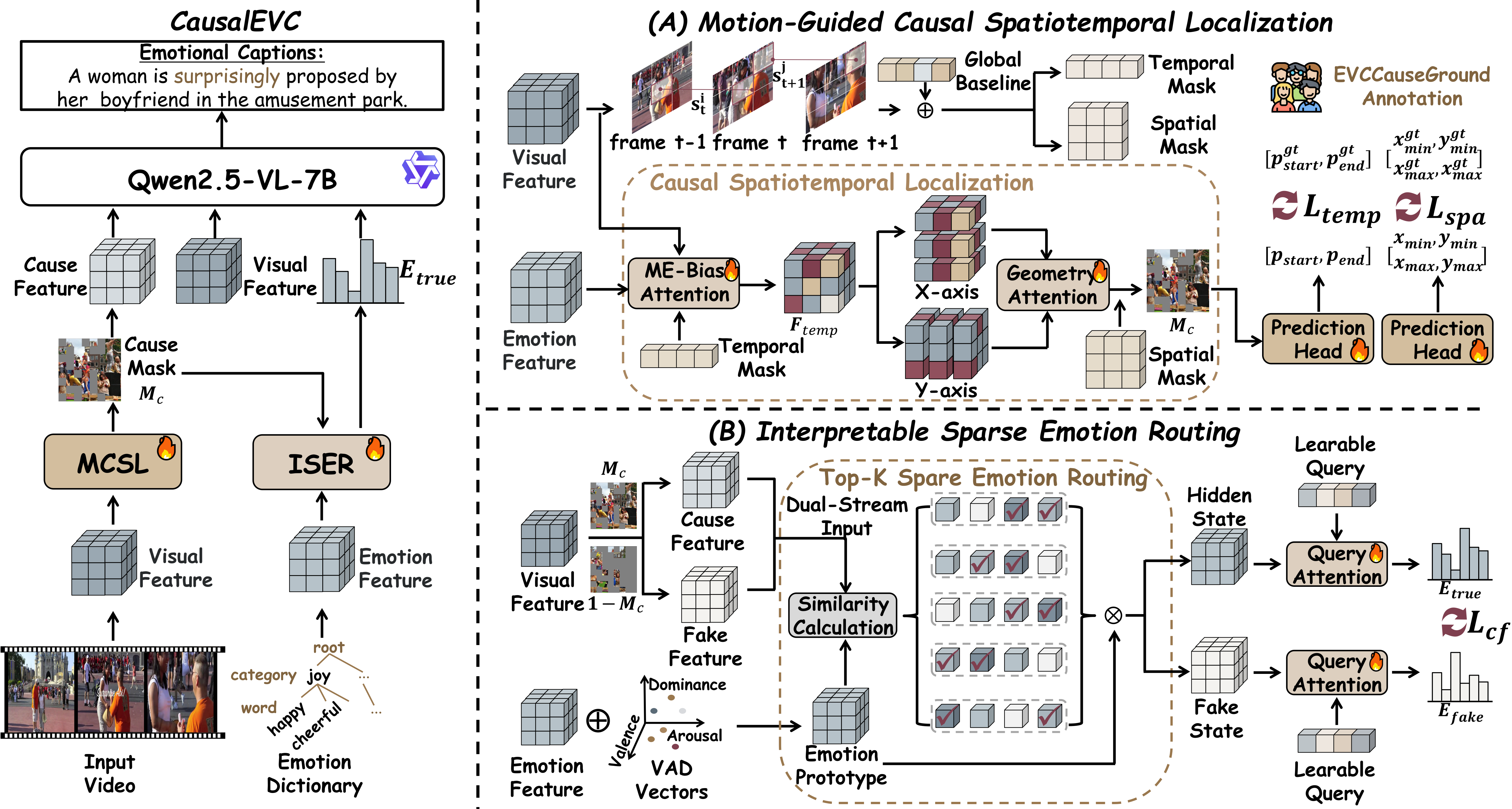}}
\caption*{Figure 4: Overview of Causal-EVC. Given the video and emotion dictionary, we perform (A) Motion-Guided Causal Spatiotemporal Localization to extract visual cause masks, and (B) Interpretable Sparse Emotion Routing to mine visual emotion cues for cause-grounding captioning.}
\label{fig2}
\vspace{-10pt}
\end{figure*}

\textbf{Evaluation Metrics.}
To effectively evaluate the performance of cause grounding, we design a cascaded metric {Spatio-Temporal Grounding Accuracy ($\text{ST-IoU}$).} Specifically, we first compute the 1D temporal $\text{IoU}$ between the predicted $[t_{start}^{pred},t_{end}^{pred}]$ and the ground-truth span. If ($\text{tIoU}=0$), the overall score is $0$. If $tIoU>0$, we calculate the 2D spatial IoU ($sIoU$) for each selected frame $t_i\in T_{interest}$. The final $\text{ST-IoU}$ is formulated as:
\begin{equation}
\text{ST-IoU} = \text{tIoU} \times \left( \frac{1}{|T_{\text{intersect}}|} \sum_{t_i \in T_{\text{intersect}}} \text{sIoU}_i \right),
\end{equation}
we report the ${\rm mIoU}$, $R@0.3$, and $R@0.5$ to robustly measure causal localization performance.

\section{Causal-EVC: A Cause-Grounding Video Captioning Framework}
In this section, we describe the Causal-EVC framework for the cause-grounding video captioning task shown in Fig. 4. Given the input video and emotion category, we first encode them to obtain the semantic features. Following the setting of previous works ~\citep{song2023emotion, ye2024dual}, we firstly down-sample the video to obtain a frame sequence $\{f_1,f_2…,f_T\}$, where $T$ denotes the number of frames. Subsequently, we leverage the pre-trained vision encoder of Qwen-2.5-VL~\citep{bai2025qwen25vltechnicalreport} to extract visual features:
\begin{equation}
    \mathcal{V}= {\rm VisEncoder([f_1,f_2,…,f_T])\in\mathbb{R}^{T \times P \times D}},
\end{equation}
{where ${\rm VideoEncoder}$ denotes the visual feature projector and $D$ is the feature dimension. $P$ is the number of patches since each frame is divided into $P$ patches.} For the emotion category, we leverage a pre-defined psychology emotional vocabulary dictionary $\Omega=\{w_i\}_{i=1}^{N_w}$, where $w_i$ denotes the $i$-th emotion word, such as ``happy'', and $N_w$ denotes the number of emotion words. Then, we leverage the pre-trained Transformer-based emotion embedding, \emph{i.e.,} GloVe ~\citep{pennington2014glove}, to encode these emotion words to ${\mathcal{E}}\in \mathbb{R}^{N_w\times D}$. The whole process can be formalized as:
\begin{equation}
    \mathcal{E} = {\rm EmoEncoder}(\Omega)\in \mathbb{R}^{N_w\times D}.\\
\end{equation}
These visual and emotional features contain rich semantic information and help to mine accurate emotional causes from video contents.

{\subsection{Motion-Guided Causal Spatiotemporal Localization}}\label{3.2}
\textbf{Motion Prior Generation.} In video streams, genuine emotional triggers are rarely isolated static objects or a specific token. Instead, they are inherently embedded in the dynamic action states and state transitions. Therefore, we propose to leverage action semantics to serve as a kinematic prior, explicitly collaborating causal localization across both temporal and spatial dimensions. To capture motion prior, we propose a pixel-level motion correlation calculation module, which quantifies motion information through calculating the similarity of all pixel positions between adjacent frames. For the $j$-th patch of the $t$-th frame $v_t^j$, the motion information is quantified as $s_t^j = 1-\frac{v_{t-1}^{j}\cdot v_t^j}{||v_{t-1}^{j}||||v_t^{j}||}$, where $s_t^j\in [0,2]$ denotes the degree of motion change. The lower the similarity of corresponding patches between adjacent frames, the greater the degree of motion change. Meanwhile, in order to eliminate biases caused by some unexpected disturbances (\emph{i.e.,}camera shake or perspective change), we calculate the global motion baseline through the average of all pixels. The motion information comprehensively considers the global motion baseline and its own motion mask matrix. Finally, we leverage the motion information to generate motion priors:
\begin{equation}
    \overline{s} = \frac{1}{T\times P}\sum_{i,j}{s}^j_t,m_t^j = {\rm ReLU}((1-\lambda_m){s}^j_t+\lambda_m \overline{s}),
\end{equation}
where $\lambda_m$ is a hyper-parameter to control the weight of the global motion baseline. By aggregating all $m_t^j$, we obtain an accurate motion prior matrix $M\in \mathbb{R}^{T\times P}$. We decouple the motion prior into temporal and spatial dimensions to achieve coordinated guidance:
\begin{equation}
    M_{temp} = \text{Softmax}(\{\sum_{j=1}^P m_t^j\}_{t=1}^T) \in \mathbb{R}^T,
    (M_{spa})_t^j = \frac{m_t^j}{\max_{k \in \{1..P\}} m_t^k + \epsilon},
\end{equation}
$M_{temp}$ indicates the peaks of the action and $M_{spa}$ represents moving regions in each frame. These constitute a powerful prior that guides two mutually synergistic perspectives on causal exploration.

\textbf{Temporal Causal Aggregation.} To achieve temporal causal aggregation, we introduce a motion-energy bias into the cross-attention mechanism to remove the impact of irrelevant temporal frames:
\begin{align}
    &Q_\tau = W_q \mathcal{E}, \quad K_\tau = W_k \mathcal{V}, \quad V_\tau = W_v \mathcal{V},\\
    A_\tau = \text{Softmax}&\left( \frac{Q_\tau K_\tau^T}{\sqrt{d}} + \alpha \cdot \text{Diag}(\log M_{temp})\right), \mathcal{F}_{temp} = A_\tau V_\tau \in \mathbb{R}^{T \times P \times D},
\end{align}
where $\alpha$ is the scaling parameter and $\text{Diag}$ transforms the 1-D curve into a diagonal matrix. The bias term significantly amplifies the weight of frames representing the peak of the movement, thereby focusing more on the emotional climax associated with intense action.

\textbf{Spatial Causal Localization.} Subsequently, we further consider key regions of the emotional causes. We calculate cross-attention along different geometric axes to capture entity boundaries:
\begin{align}
    Q_s^g = W_{s,q}^g \mathcal{E}, \quad K_s^g = W_{s,k}^g \mathcal{F}_{temp}, \quad V_s^g = W_{s,v}^g \mathcal{F}_{temp},\\
    A_s^g = \text{Softmax}(Q_s^g K_s^{gT}), \mathcal{O}_s = A_t^y(A_t^x V_s^g)^T\in \mathbb{R}^{T \times P \times D},
\end{align}
where $g\in\{x,y\}$ represents the horizontal and vertical axes. $\mathcal{O}_s$ highlights all salient emotion-related regions in the scene. Finally, we utilize $M_{spa}$ to modulate $\mathcal{O}_s$ and obtain causal features and their corresponding masks:
\begin{equation}
    \mathcal{F}_{c} = \mathcal{O}_s \odot M_{spa} \in \mathbb{R}^{T \times P \times D},
    {M}_c = \sigma(\text{Conv3D}(\mathcal{F}_{c})) \in [0, 1]^{T \times P \times 1},
\end{equation}
where $\sigma$ is the Sigmoid activation function. To precisely ground the spatiotemporal labels, we design two MLP heads and objective functions to achieve powerful supervision. Instead of predicting frame-wise and pixel-wise probabilities, we regress the normalized temporal indexes and spatial coordinates. First, we aggregate $\mathcal{F}_{c}$ to obtain a spatial visual vector $f_s\in \mathbb{R}^{T\times D}$ and a compact visual vector $f_{g} \in \mathbb{R}^D$:
\begin{equation}
    f_s=\text{AvgPool}_{P}(\mathcal{F}_{c}),f_{g} = \text{MaxPool}_{T}(\text{Conv1D}(f_s)),
\end{equation}
Then, temporal and spatial heads are used to predict the temporal indexes and spatial coordinates:
\begin{equation}
    ([\hat{t}_{start}, \hat{t}_{end}],[\hat{x}_{min}^{(t)}, \hat{y}_{min}^{(t)}, \hat{x}_{max}^{(t)}, \hat{y}_{max}^{(t)}]) = (\sigma(\text{MLP}_{temp}(f_{g})),
     \sigma(\text{MLP}_{spa}(f_{s,t}))),
\end{equation}
where $\sigma$ is the Sigmoid activation function and $\text{MLP}$ is the multi-layer projection.

\textbf{Causal Grounding Objective.} During training, given the ground-truth temporal span $[t^{gt}_{start}, t^{gt}_{end}]$, we apply the SmoothL1 loss~\citep{huber1992robust} to robustly supervise the boundary regression:
\begin{equation}
    \mathcal{L}_{temp} = \text{SmoothL1}(\hat{p}_{start}, p^{gt}_{start}) + \text{SmoothL1}(\hat{p}_{end}, p^{gt}_{end}),
\end{equation}
For the supervision of spatial regions, denoting the predicted and ground-truth bounding box as $B^{pred}_t$ and $B^{gt}_t$, we employ a combination of SmoothL1 loss and Generalized IoU loss~\citep{hamid2019giou}, which effectively penalizes scale mismatches and prevents gradient vanishing when there is no overlap between boxes:
\begin{align}
    \mathcal{L}_{GIoU}(B^{pred}_t, B^{gt}_t) = 1 - \left( \text{IoU}(B^{pred}_t,B^{gt}_t) - \frac{\text{Area}(C_t) - \text{Area}(B^{pred}_t \cup B^{gt}_t)}{\text{Area}(C_t)} \right),\\
    \mathcal{L}_{spa} = \frac{1}{\sum_{t=1}^T \mathbb{I}^{gt}_t} \sum_{t=1}^T \mathbb{I}^{gt}_t \cdot \Big[ \lambda_{L1} \mathcal{L}_{L1}(B^{pred}_t, B^{gt}_t) + \lambda_{giou} \mathcal{L}_{GIoU}(B^{pred}_t, B^{gt}_t) \Big],
\end{align}
where $C_t$ is the minimum enclosing box containing both $B^{pred}_t$ and $B^{gt}_t$, $\lambda_{L1}$ and $\lambda_{giou}$ are balancing weights. $\mathbb{I}^{gt} \in \{0, 1\}^T$ is a temporal indicator mask to ensure only calculating the spatial loss on frames where the genuine causal action occurs.

{\subsection{Interpretable Sparse Emotion Routing}}\label{3.3}
\textbf{Emotion Prototype Augmentation.} Unlike objective elements, the semantic information of emotion words is characterized by complexity and subjectivity. Relying on a unified text encoder for emotional mining is insufficient. Therefore, we introduce the VAD lexicon~\citep{mohammad2018obtaining} ${R}=\{(v_i,a_i,d_i)\}_{i=1}^{N_w}\in\mathbb{R}^{N_w\times 3}$, where each entry is normalized to $[-1,1]$ (and out-of-lexicon words are set to $\mathbf{0}$), to obtain an interpretable emotion-augmented representation. Given the emotion vocabulary $\mathcal{E}$ and the VAD lexicon, we stack these triplets to obtain the emotion prototype features:
\begin{equation}
\hat{\mathcal{E}}= {\rm LayerNorm}(\mathcal{E} \oplus W_eR) \in \mathbb{R}^{N_w\times D},
\end{equation}
The $\hat{\mathcal{E}}$ contains both textual semantic and psychological attributes, enhancing the interpretability and facilitating the extraction of emotional semantics from visual information.

\textbf{Top-K Sparse Emotion Routing.} Affective semantics are characterized by significant vagueness and ambiguity. Traditional methods are highly susceptible to interference from irrelevant emotional information. Therefore, we design a sparse emotion routing mechanism. First, instead of using redundant visual features, we employ the cause mask $M_c$ to filter and obtain visual causal features $\mathcal{C}=V\odot M_c$, which are then used to query emotion prototypes. Subsequently, to prevent the traditional attention mechanism from capturing ambiguous emotional semantics, we enforce matching only the top-$k$ emotion categories. Specifically, we first compute the similarity matrix $A^e$ between $\mathcal{C}$ and $\hat{\mathcal{E}}$. Subsequently, for each visual token, we retain only the top-$K$ values and set the rest to $-\infty$, thereby obtaining the sparse attention matrix $\hat{A}_e$ to route the emotion extraction:
\begin{equation}
\hat{A}^e_{i, j} =
\begin{cases}
\frac{\exp(A^e_{i,j})}{\sum_{m \in \text{Top-K}(A^e_i)} \exp(A_{i,m})}, & \text{if } j \in \text{Top-K}(A^e_i) \\
0, & \text{otherwise}
\end{cases},\mathcal{H}_{true} = \hat{A}^e (\hat{\mathcal{E}}W_h)\in\mathbb{R}^{T\times D},
\end{equation}
$\mathcal{H}_{true}$ denotes the visual hidden states related to the emotions. Finally, we leverage a set of learnable category query vectors $\mathbf{q}_{emo}\in \mathbb{R}^{N_w\times D}$ to mine emotions:
\begin{equation}
\hat{\mathcal{E}}_{true} = \text{MHCA}(\text{Query}=\mathbf{q}_{emo}, \text{Key}=\mathcal{H}_{true}, \text{Value}=\mathcal{H}_{true}) \in \mathbb{R}^{N_w \times D},
\end{equation}
besides, to achieve the accurate supervision for emotion mining, we introduce a counterfactual causal alignment module. Specifically, we replace the visual causal features $\mathcal{C}$ in the aforementioned process with counterfactual visual features $\hat{\mathcal{C}}=V\odot (1-M_c)$ to obtain counterfactual emotion features $\hat{\mathcal{E}}_{fake}$. Then, we calculate the cosine similarity between the ground-truth and the predicted emotion features to obtain the emotion logits:
\begin{equation}
    z_{true} = {\rm Cos}({\mathcal{E}_{gt}}
    ,\hat{\mathcal{E}}_{true})\in\mathbb{R}^{N_w},z_{fake} = {\rm Cos}(\mathcal{E}_{gt},\hat{\mathcal{E}}_{fake})\in\mathbb{R}^{N_w},
\end{equation}
where $\mathcal{E}_{gt}$ denotes the prototype embedding of the ground-truth emotion labels. Then, we employ an NCE loss~\citep{fu2024linguistic,radford2021learning,liu2026continual,li2026cspcl} on $(z_{true},z_{fake})$ pair to guide the model to learn accurate emotion features from genuine causal features, rather than relying on shortcut learning from spurious visual backgrounds.
\begin{equation}
\mathcal{L}_{cf} = - \log \frac{\exp(z_{true}^{(y)} / \tau_{cf})}{\exp(z_{true}^{(y)} / \tau_{cf}) + \exp(z_{fake}^{(y)} / \tau_{cf})},
\end{equation}
where $\tau_{cf}$ is the temperature parameter. After obtaining sufficient semantic information including video, cause, and emotion features, we concatenate them to generate a unified multi-modal semantic representation $M=[\mathcal{V};\mathcal{C};\mathcal{E}_{true}]\in\mathbb{R}^{(T\times 2P+N_w)\times D}$. Finally, we send them into the decoder with Qwen-2.5-VL model~\citep{bai2025qwen25vltechnicalreport} to generate the emotional captions.

% \textbf{Optimization Objective.} In addition to the causal grounding and counterfactual alignment loss, we also introduce two emotion-aware optimization objectives. Firstly, since the important role of emotional words in caption generation, we use emotion-focused cross-entropy loss that adds a penalty term on emotional words. Furthermore, to provide more sufficient emotional loss, we also design an emotional classification loss by adding a simple classification header for $\mathcal{E}_{true}$ to compute the classification loss:
% \begin{equation}
%     \mathcal{L}_{e}=\begin{cases}-(1+\theta)\sum_{t=1}^{N_T}\log P(y_{t}|y_{<t}),&{\rm if}\ y_{t}\in \Omega,\\\\-\sum_{t=1}^{N_T}\log P(y_{t}|y_{<t}),&{\rm otherwise},\end{cases},\mathcal{L}_{cls}=-\sum_{e \in \Omega}\log P(e|d),
% \end{equation}
% where $\theta$ is a hyper-parameter that controls the level of punishment in the equation when $y_t$ is an emotional word, $N_T$ denotes the sequence length, and $d$ is the emotional probability distribution. The overall loss is the weighted sum of five losses:
% \begin{equation}
% \mathcal{L}
% =\lambda_{e}\mathcal{L}_{e}
% +\lambda_{cls}\mathcal{L}_{cls}
% +\lambda_{temp}\mathcal{L}_{temp}
% +\lambda_{spa}\mathcal{L}_{spa}
% +\lambda_{cf}\mathcal{L}_{cf},
% \end{equation}
% where $\lambda_{e}$, $\lambda_{cls}$, $\lambda_{temp}$, $\lambda_{spa}$, and $\lambda_{cf}$ are hyper-parameters that balance the four objectives. Our model is trained end-to-end by minimizing the overall loss $\mathcal{L}$.
\begin{table}[t]
\centering
\caption{\label{main}{Performance for emotional video captioning task on two benchmarks. The best results are highlighted in bold.}}
\scalebox{0.72}{
\begin{tabular}{l|cc|ccccc|cc}
\toprule
\multirow{2}{*}{\textbf{Methods}}  &  \multicolumn{2}{c|}{\textbf{Emotion}} &\multicolumn{5}{c|}{\textbf{Semantic}} & \multicolumn{2}{c}{\textbf{Hybrid}}\\
~ &${\rm Acc}_{sw}$$\uparrow$& ${\rm Acc}_c$$\uparrow$&{BLEU-1$\uparrow$}   & {BLEU-4$\uparrow$}   & {METEOR$\uparrow$} & {ROUGE$\uparrow$} & {CIDEr$\uparrow$}&BFS$\uparrow$&CFS$\uparrow$  \\ 
\toprule
\multicolumn{10}{c}{\textbf{EVC-MSVD}}\\
\cmidrule{1-10}
\rowcolor{gray!15}{SA-LSTM} ~\citeyearpar{wang2018reconstruction}  &68.8 &67.2 &80.7 &45.5 &33.0 &68.2 &72.1 &59.0 &71.3\\
FT        ~\citeyearpar{wang2021emotion}        &69.4 &67.1 &77.2  &36.3 &29.0 &63.4 &62.5 &52.5 &63.7\\
SGN       ~\citeyearpar{ryu2021semantic}        &73.9 &73.1 &77.5 &41.1 &30.6 &63.6 &71.0 &56.4 &71.5\\
CANet     ~\citeyearpar{song2022contextual}     &78.7 &76.8 &78.5 &41.8 &30.8 &65.7 &74.4 &57.9 &75.1\\
{VEIN}    ~\citeyearpar{song2024emotional}     &82.7 &82.1 &82.0 &45.9 &33.0 &69.0 &79.6 &62.4 &80.2\\
{EPAN}    ~\citeyearpar{song2023emotion}       &84.1 &82.8 &82.5 &46.2 &34.4 &69.8 &80.6 &63.1 &81.1\\
{DCGN}    ~\citeyearpar{ye2024dual}            &${86.5}$  &${85.7}$& ${84.5}$ & ${48.7}$ & ${35.7}$&${71.0 }$   & ${85.2}$  & ${65.7}$&${86.6}$ \\
% \cmidrule{1-14}
GLM-EER~\citeyearpar{wang2026adaptive}&88.2& 86.6 &92.2 &68.3 &43.8& 80.1 &153.9 &78.8 &140.6\\
{MM-ECPE}~\citeyearpar{ye2025multi}&{{90.4}} &{89.1} & \underline{96.9}	&\textbf{71.4}	&\underline{45.5}	&\underline{83.1}	&\underline{168.3} &\underline{81.8}	&\underline{152.6} \\
SAGML~\citeyearpar{ma2026glm}&\underline{93.5}& \underline{92.3} &93.6 &53.8 &40.8 &77.2& 96.1 &71.5 &95.5\\
\rowcolor{oursburgundy}\textbf{Causal-EVC(Ours)}&\textbf{94.2} &\textbf{92.9} &\textbf {97.4}	&\underline{71.2}	&\textbf{46.2}	&\textbf{84.6}	&\textbf{169.8} &\textbf{82.7}	&\textbf{154.6} \\
\toprule
\multicolumn{10}{c}{\textbf{EVC-VE}}\\
\cmidrule{1-10}
\rowcolor{gray!15}{SA-LSTM} ~\citeyearpar{wang2018reconstruction} & 48.6 &47.1& 71.0 &22.5 & 19.6&40.7 &30.2& 38.9& 33.7 \\
CANet ~\citeyearpar{song2022contextual} &41.9 &39.7 &66.9 &19.3 &18.2 &37.9 &23.3 &33.9 &26.8\\
{VEIN} ~\citeyearpar{song2024emotional}  & 57.4 &56.8&71.6&26.3&20.9&41.7&33.4&43.0& 39.2\\
{EPAN} ~\citeyearpar{song2023emotion}  & 63.8 &62.3& 73.6 &27.0 & 21.2&42.3 &34.7&45.0 &40.4\\
{DCGN} ~\citeyearpar{ye2024dual}  & 	${71.0}$  &${69.4}$& ${74.5}$  & ${28.1}$ & ${23.4}$&${47.7}$    &${41.5}$& ${47.3}$&${46.9}$  \\	
% \cmidrule{1-14}
{HEART}~\citeyearpar{han2026heart} & 55.8 &54.2 &71.5 &22.7 &19.7 &41.2 &28.8 &40.3 &34.0\\
{MM-ECPE}~\citeyearpar{ye2025multi} & {73.4}	&{72.3}& {76.8}	&{28.9}	&{24.7}	&{49.5}	&{65.2}	 &{49.2}	&{66.7}	  \\
SAGML~\citeyearpar{ma2026glm}&\underline{80.2} &\underline{79.6} &\underline{82.7} &\underline{31.8} &\underline{26.9}& \underline{52.7}& \underline{71.8} &\underline{53.8} &\underline{73.4}\\
\rowcolor{oursburgundy}\textbf{Causal-EVC(Ours)}&{\textbf{82.8}} &\textbf{81.5} & \textbf{85.1}	&\textbf{32.4}	&\textbf{27.7}	&\textbf{54.6}	&\textbf{74.0} &\textbf{55.3}	&\textbf{75.6} \\
% \toprule
% \multicolumn{10}{c}{\textbf{EVC-Combine}}\\
% \cmidrule{1-10}
% {SA-LSTM} ~\citep{wang2018reconstruction} &53.4 &50.7& 70.6& 25.4 & 21.0&45.9& 38.8& 41.2 &41.5 \\
% FT ~\citep{wang2021emotion}&51.2 &49.6& 67.6  &21.6 &20.4 &43.1 &29.0& 37.6 &33.3\\
% SGN~\citep{ryu2021semantic}&50.4&48.6&68.7&24.0&20.1&44.8&35.5&39.1&38.3\\
% CANet ~\citep{song2022contextual} &53.7 &52.7 &68.1 &22.5 &19.7 &43.7 &34.5 &38.8 &38.2\\
% {VEIN} ~\citep{song2024emotional}& 59.0 &57.6&72.1&27.1&21.6&46.8&39.4&43.6& 43.1\\
% {EPAN} ~\citep{song2023emotion} & 69.3 &67.2& 74.4  &28.0 & 23.0&47.1& 43.0&47.0 &48.0\\
% {DCGN} ~\citep{ye2024dual}  & ${74.8}$ &${73.1}$ & ${75.6}$ & ${28.5}$ & ${24.9}$&${51.7}$    & ${49.8}$& ${48.5}$ &${51.7}$  \\
% HEART~\citep{han2026heart}&66.2 &64.1 &74.8  &27.8& 23.3& 49.8& 37.7 &46.3& 43.7\\
% GLM-EER~\citep{wang2026adaptive}&64.9 &63.2 &75.4&30.0 &26.1 &51.1& 64.5 &47.6 &64.4\\
% {MM-ECPE}~\citep{ye2025multi} & {75.6}&{73.8}&{78.1}	&{30.2}	&{26.4}	&{53.8}	&{67.9}	&{50.4}	&{69.3} \\
% SAGML~\citep{ma2026glm}&82.3 &81.5 &85.7 &34.8 &28.5 &57.2& 76.9& 55.7 &77.9\\
% % \cmidrule{1-14}
% \textbf{Ours} &{\textbf{84.4}} &\textbf{82.6} & \textbf{86.9}	&\textbf{35.5}	&\textbf{29.1}	&\textbf{57.7}	&\textbf{78.8} &\textbf{56.9}	&\textbf{79.9} \\
\bottomrule
\end{tabular}}
\vspace{-15pt}
\end{table}

\section{Results and Discussion}

\subsection{Main Results}

\begin{wrapfigure}{r}{0.55\textwidth} % r 表示居右，占 0.5 宽度
  \centering
      \setlength{\aboverulesep}{0pt}
\setlength{\belowrulesep}{0pt}
  \vspace{-10pt} % 根据需要调整表格上方的间距
\caption*{Table 2: Results on EVC-CauseGround.}
\label{tab:results_causal_faithfulness}
\scalebox{0.8}{\begin{tabular}{l|ccc|c}
\toprule
Methods & $mIoU$ & $R@0.3$ & $R@0.5$ & ${\rm Acc_{sw}}$ \\ 
\midrule
\rowcolor{gray!15}MM-ECPE~\citeyearpar{ye2025multi} &18.6 &28.7 &12.6 &53.1 \\
\midrule
Momentor~\citeyearpar{qian2024momentor} &21.5 &35.8 &20.4 &58.5 \\
VTimeLLM~\citeyearpar{huang2024vtimellm} &23.7 &42.1 &22.0 &61.4 \\
VideoChat-T~\citeyearpar{zeng2025timesuite} &37.6 &61.9 &40.8 &78.4 \\
\rowcolor{gray!15}Qwen2.5-VL-7B~\citeyearpar{bai2025qwen25vltechnicalreport} &\underline{39.8} &\underline{62.5} &\underline{41.3} &\underline{80.8} \\
\midrule
Gemini-2.5-Flash~\citeyearpar{comanici2025gemini} &30.7 &50.9 &27.1 &67.0 \\
GPT-4o~\citeyearpar{hurst2024gpt} &35.1 &55.5 &33.2 &73.9 \\
\rowcolor{oursburgundy}\textbf{Ours} &\textbf{46.7} &\textbf{70.4} &\textbf{49.2} &\textbf{90.3} \\
\bottomrule
\end{tabular}}
\vspace{-10pt}
\end{wrapfigure}

As shown in Table \ref{main}, Causal-EVC consistently achieves the best performance across almost all metrics, \emph{i.e.,} improving +4.2\% with $Acc_{sw}$ than MM-ECPE and +15.7\% with BFS than SAGML on EVC-MSVD. More notably, on the larger EVC-VE benchmark, Causal-EVC also outperforms SAGML, \emph{i.e.,} +2.4\%/+3.0\% improvements with $Acc_{c}$/BLEU-4. These prove that Causal-EVC effectively achieves better emotional description generation. Besides, Table 2 reports the cause-grounding performance on EVC-CauseGround. Causal-EVC demonstrates substantial dominance across all metrics. Compared to specialized visual grounding models, our model leads +67.2\%/13.7\% than VTimeLLM~\citep{huang2024vtimellm}/VideoChat-T~\citep{zeng2025timesuite} on $R@0.3$. While these models excel at objective localization, they are prone to hallucinations when it comes to emotional visual attribution due to their lack of specific emotion understanding components. Besides, compared to general MLLMs, our model leads +52.1\%/33.0\% than Gemini-2.5-Flash~\citep{comanici2025gemini}/GPT-4o~\citep{hurst2024gpt} on ${\rm mIoU}$. Despite the powerful comprehension capabilities of these models, the dual requirements of emotional understanding and visual grounding exacerbate the frequency of hallucinations, which demonstrates that Causal-EVC improves the authentic and robust emotion-cause reasoning and alleviates the co-occurrence bias.

\begin{table*}[t]
\centering
% ======================== 第二行：左侧表格 ========================
\begin{minipage}[t]{0.48\textwidth}
\centering
\caption*{Table 3: The ablation study for dense MCSL.}
\label{tab:ablation_dense_components}
\vspace{2pt}
\scalebox{0.7}{
\begin{tabular}{c|cc|cccc}
\toprule
\multicolumn{3}{c|}{Components} & \multirow{2}*{${\rm Acc}_{sw}$} & \multirow{2}*{BLEU-4} & \multirow{2}*{CIDEr} & \multirow{2}*{CFS} \\
Motion & Temporal & Spatial & & & & \\ 
\midrule
\multirow{4}{*}{$\times$} & \cellcolor{gray!15}$\times$ & \cellcolor{gray!15}$\times$ &\cellcolor{gray!15}92.9 &\cellcolor{gray!15}67.8 &\cellcolor{gray!15}154.4 &\cellcolor{gray!15}142.0 \\ 
~ & $\checkmark$ & $\times$ &89.4 &68.6 &156.9 &143.2 \\ 
~ & $\times$ & $\checkmark$ &91.2 &69.7 &158.5 &144.9 \\ 
~ & \cellcolor{oursburgundy}$\checkmark$ & \cellcolor{oursburgundy}$\checkmark$ &\cellcolor{oursburgundy}90.8 &\cellcolor{oursburgundy}70.3 &\cellcolor{oursburgundy}160.2 &\cellcolor{oursburgundy}146.2 \\ 
\midrule
\multirow{4}{*}{$\checkmark$} & \cellcolor{gray!15}$\times$ & \cellcolor{gray!15}$\times$ &\cellcolor{gray!15}90.5 &\cellcolor{gray!15}67.2 &\cellcolor{gray!15}152.8 &\cellcolor{gray!15}140.2 \\ 
~ & $\checkmark$ & $\times$ &93.4 &70.1 &160.6 &147.0 \\ 
~ & $\times$ & $\checkmark$ &93.7 &70.6 &163.9 &149.7 \\ 
~ & \cellcolor{oursburgundy}$\checkmark$ & \cellcolor{oursburgundy}$\checkmark$ & \cellcolor{oursburgundy}\textbf{94.2} & \cellcolor{oursburgundy}\textbf{71.2} & \cellcolor{oursburgundy}\textbf{169.8} & \cellcolor{oursburgundy}\textbf{154.6} \\
\bottomrule
\end{tabular}
}
\end{minipage}
\hfill
% ======================== 第二行：右侧表格 ========================
\begin{minipage}[t]{0.48\textwidth}
\centering
\caption*{Table 4: The ablation study for proposed losses.}
\label{tab:ablation_losses}
\vspace{2pt}
\scalebox{0.7}{
\begin{tabular}{c|cc|cccc}
\toprule
\multicolumn{3}{c|}{Components} & \multirow{2}*{${\rm Acc}_{sw}$} & \multirow{2}*{BLEU-4} & \multirow{2}*{CIDEr} & \multirow{2}*{CFS} \\
$\mathcal{L}_{cf}$ & $\mathcal{L}_{temp}$ & $\mathcal{L}_{spa}$ & & & & \\ 
\midrule
\multirow{4}{*}{$\times$} & \cellcolor{gray!15}$\times$ & \cellcolor{gray!15}$\times$ &\cellcolor{gray!15}87.3 &\cellcolor{gray!15}66.7 &\cellcolor{gray!15}153.2 &\cellcolor{gray!15}139.9 \\ 
~ & $\checkmark$ & $\times$ &87.8 &68.5 &157.4 &143.4 \\ 
~ & $\times$ & $\checkmark$ &87.4 &68.9 &158.3 &144.0 \\ 
~ & \cellcolor{oursburgundy}$\checkmark$ & \cellcolor{oursburgundy}$\checkmark$ &\cellcolor{oursburgundy}88.4 &\cellcolor{oursburgundy}70.1 &\cellcolor{oursburgundy}164.0 &\cellcolor{oursburgundy}148.8 \\ 
\midrule
\multirow{4}{*}{$\checkmark$} & \cellcolor{gray!15}$\times$ & \cellcolor{gray!15}$\times$ &\cellcolor{gray!15}90.8 &\cellcolor{gray!15}67.2 &\cellcolor{gray!15}155.9 &\cellcolor{gray!15}142.8 \\ 
~ & $\checkmark$ & $\times$ &92.1 &69.0 &159.3 &145.8 \\ 
~ & $\times$ & $\checkmark$ &91.6 &69.4 &160.2 &146.4 \\ 
~ & \cellcolor{oursburgundy}$\checkmark$ & \cellcolor{oursburgundy}$\checkmark$ & \cellcolor{oursburgundy}\textbf{94.2} & \cellcolor{oursburgundy}\textbf{71.2} & \cellcolor{oursburgundy}\textbf{169.8} & \cellcolor{oursburgundy}\textbf{154.6} \\
\bottomrule
\end{tabular}
}
\end{minipage}
\vspace{-15pt}
\end{table*}

\subsection{Ablation Studies}

\begin{wrapfigure}{r}{0.5\textwidth} % r 表示居右，占 0.5 宽度
  \centering
      \setlength{\aboverulesep}{0pt}
\setlength{\belowrulesep}{0pt}
  \vspace{-10pt} % 根据需要调整表格上方的间距
\caption*{Table 5: The ablation study for proposed modules.}
\label{tab:ablation_mcsl_iser}
\scalebox{0.75}{\begin{tabular}{cc|cccc}
\toprule
\multicolumn{2}{c|}{Components} & \multirow{2}*{${\rm Acc}_{sw}$} & \multirow{2}*{BLEU-4} & \multirow{2}*{CIDEr} & \multirow{2}*{CFS} \\
MCSL & ISER & & & & \\
\midrule
\rowcolor{gray!15}$\times$ & $\times$ &82.1 &65.3 & 150.9& 137.0\\
$\checkmark$ & $\times$ &85.7 &70.2 &165.4 &149.4 \\ 
$\times$ & $\checkmark$ &92.9 &67.8 & 154.4&142.0 \\ 
\rowcolor{oursburgundy}$\checkmark$ & $\checkmark$ & \textbf{94.2} & \textbf{71.2} & \textbf{169.8} & \textbf{154.6} \\
\bottomrule
\end{tabular}}
\vspace{-15pt}
\end{wrapfigure}

\textbf{Discussion on proposed modules.} As shown in Table 5, we make an ablation study on proposed modules. We observe that each module could improve its performance independently. Specifically, MCSL effectively localizes the cause cues, which helps to generate more accurate captions for crucial emotion-related events, resulting in obvious improvements on semantic metrics, \emph{i.e.,} +7.5\%/+9.1\% improvements on BLEU-4/CFS. Besides, ISER effectively utilizes causal masks and counterfactual interventions to achieve accurate emotion cue extraction, resulting in obvious improvements in emotion accuracy, \emph{i.e.,} +13.2\% improvements on ${\rm Acc}_{sw}$. Finally, we combine two modules to achieve the best performance, \emph{i.e.,} +14.7\%/+12.8\% improvements on ${\rm Acc}_{sw}$/CFS.

\noindent\textbf{Discussion on MCSL.} We further discuss the dense components in MCSL module in Table 3. We first observe that solely relying on motion dynamics to generate the causal mask leads to a performance degradation. We analyze that pixel-wise differences lack semantic discriminability and are easily corrupted by environmental disturbances, such as swaying leaves. However, when motion information is used as a kinematic prior to guide both spatial and temporal localization, the model achieves a substantial performance surge. This result strongly validates that raw motion dynamics are inherently suited as guiding priors rather than standalone causes, while the genuine causal triggers fundamentally reside in the structural semantic entities executing these dynamic actions.

\begin{wrapfigure}{r}{0.5\textwidth} % r 表示居右，占 0.5 宽度
  \centering
      \setlength{\aboverulesep}{0pt}
\setlength{\belowrulesep}{0pt}
  \vspace{-10pt} % 根据需要调整表格上方的间距
  \caption*{Table 6: Results of different Top-K routing.}
  \label{topk}
  \small
\begin{tabular}{l|cccc}
\toprule
{Top-K}  &  ${\rm Acc}_{sw}$   & {BLEU-4} & {CIDEr} & CFS  \\  
\midrule
\rowcolor{gray!15}1 & 49.6    & 51.5 & 115.9 &  102.5  \\
5 &  88.6   & 67.4 & 149.8 &  137.4  \\
\rowcolor{oursburgundy}10 & 94.2    &  71.2& 169.8 &  154.6  \\
20 &  92.9   & 70.1 & 161.8 &  147.8  \\
\rowcolor{gray!15}50 &  92.2  & 68.9 & 156.3 &  143.3  \\
\bottomrule
\end{tabular}
  \vspace{-10pt} % 调整表格下方的间距
\end{wrapfigure}

\noindent\textbf{Discussion on Top-K Routing.} As shown in Table 6, we discuss the performance of different Top-K routing. First, we observe that an excessively tight constraint (i.e., $K=1$) leads to a performance collapse. This degradation occurs because forcing a single-prototype selection severely undermines the model's fault tolerance and fails to accommodate the emotional diversity nature. 
Conversely, scaling $K$ to an overly large value (e.g., $K=50$) causes trivial performance. An unconstrained routing introduces redundant and irrelevant emotion prototypes, which compromises the sparsity bottleneck and dilutes the authentic causal emotion features with background semantic noise.
Overall, we set $K=10$ as the default configuration, which strikes the optimal trade-off between affective expressiveness and noise suppression.

\noindent\textbf{Discussion on objective functions.} Besides, we conduct an ablation study to discuss the effectiveness of three specialized objective functions $\mathcal{L}_{temp}$, $\mathcal{L}_{spa}$, and $\mathcal{L}_{cf}$. As shown in Table 4, applying $\mathcal{L}_{cf}$ brings significant improvements on emotion accuracy. It isolates spurious background information by counterfactual contrastive learning, ensuring that emotion predictions stem from genuine causal triggers. Besides, applying $\mathcal{L}_{temp}$ and $\mathcal{L}_{spa}$ brings remarkable improvements on semantic metrics. They leverage ground-truth spatiotemporal tube annotations to supervise cause grounding, ensuring the accuracy of key cause-related words in emotional captions. Finally, we employ these objective functions in combination to achieve end-to-end training, ensuring the accuracy in both emotional description and causal localization.

\subsection{Qualitative Results}

\begin{figure*}[t]        
\center{\includegraphics[width=1\linewidth] {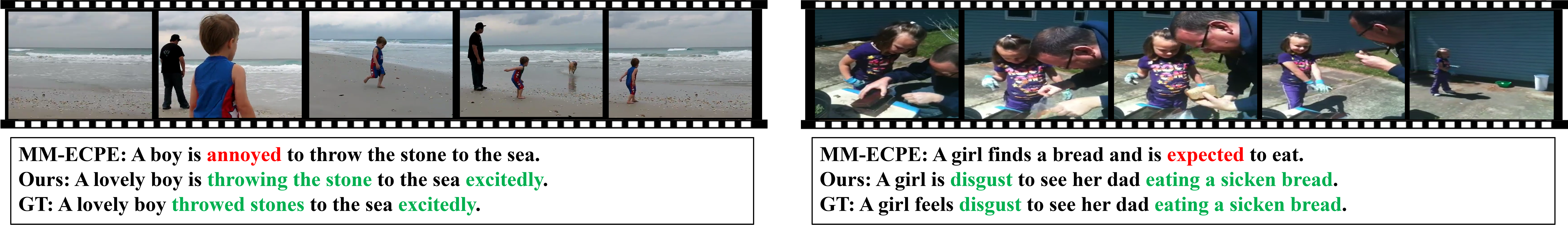}}
\caption*{Figure 5: Comparison of caption generation between MM-ECPE and our Causal-EVC.}
\label{cap}
\vspace{-15pt}
\end{figure*}

\begin{wrapfigure}{r}{0.6\textwidth} 
    \centering
    \includegraphics[width=0.6\textwidth]{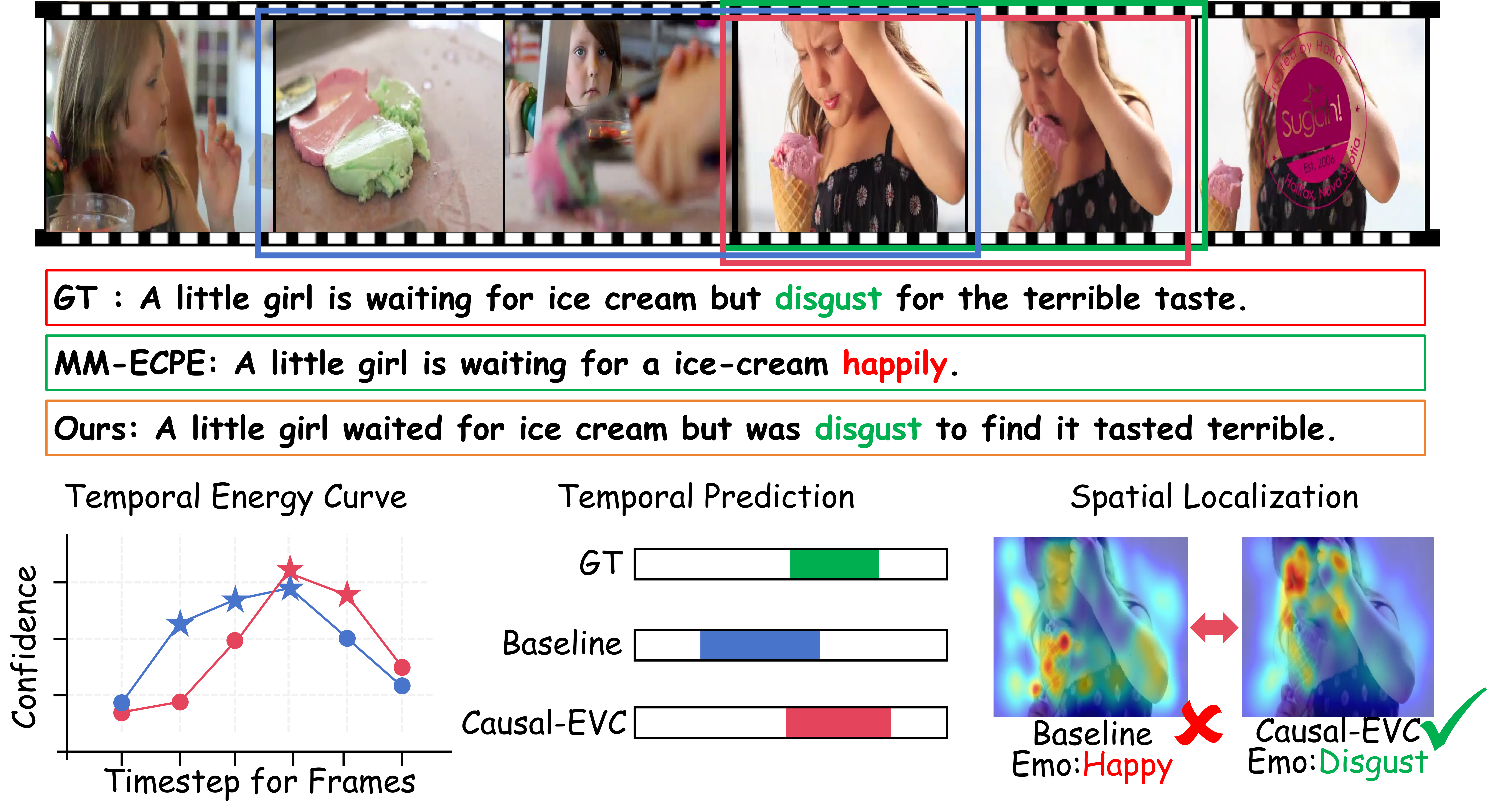} 
    \caption*{Figure 6: Case study for cause localization.}
    \label{loca}
    \vspace{-10pt}
\end{wrapfigure} 

\textbf{Case Study of Causal Localization.} We first make a comparison of caption generation between MM-ECPE and Causal-EVC shown in Fig. 5. MM-ECPE generates incorrect emotions `annoyed' and `expected', due to co-occurrence bias for `throwing stones' and insufficient localization for `sicken bread'.
Besides, we display the effect of cause localization in Fig. 6. We first observe from the Temporal Energy Curve that the baseline is easily misled by the early preparation of the ice cream, peaking prematurely. Instead, our curve surges precisely when the girl experiences a sudden sensory discomfort. Consequently, our Causal-EVC predicts a tighter temporal span that aligns with GT labels than baselines.
More crucially, from the spatial heatmaps, we observe that our model successfully eliminates the spurious correlation. Tricked by the prior bias ``ice-cream$\rightarrow$happy``, the baseline focuses predominantly on the ice cream region, mistakenly inferring a ``Happy'' emotion while entirely overlooking the child's adverse reaction. In contrast, Causal-EVC penalizes the confounder and concentrates on the disgust expressions of the girl, inferring a ``Disgust`` emotion successfully.

\begin{wrapfigure}{r}{0.6\textwidth} 
    \centering
    \includegraphics[width=0.6\textwidth]{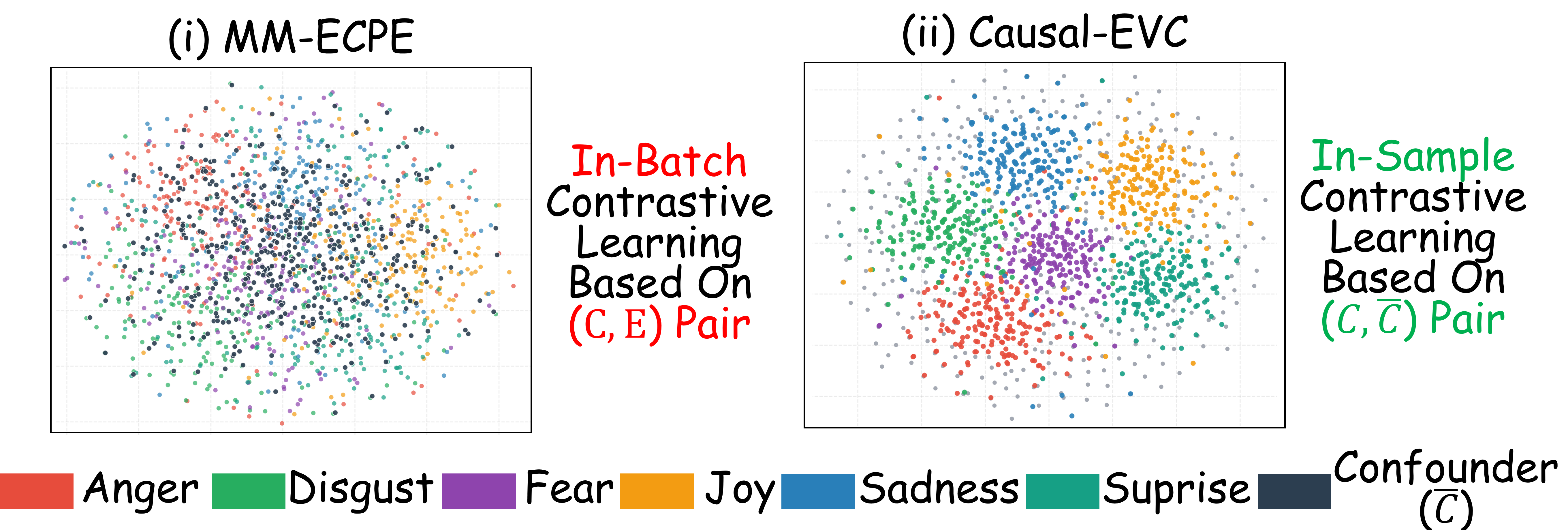} 
    \caption*{Figure 7: t-SNE visualization of cause representation.}
    \label{tsne}
    \vspace{-10pt}
\end{wrapfigure} 

\textbf{t-SNE Visualization of Cause Representation.} We visualize the t-SNE distributions of $\mathcal{C}$ across emotion spaces alongside $\hat{C}$. As shown in Fig. 7, we first observe that compared MM-ECPE, our Causal-EVC achieves better semantic clustering. This stems from the MCSL module, which synergistically leverages motion priors to steer cause localization under strict coordinate supervision.
More crucially, In MM-ECPE, $\hat{C}$ are systematically entangled within the respective emotion clusters. This confirms our findings that conventional in-batch contrastive learning on $(\mathcal{C}, \mathcal{E})$ suffers from shortcut learning, mistakenly encoding background contexts as emotional evidence. Instead, under our Causal-EVC, $\hat{C}$ are uniformly distributed across the entire latent space. This validates that our $\mathcal{L}_{cf}$ on $(C, \hat{C})$ pairs successfully severs the spurious learning from environmental contexts, strictly compelling to deduce affective semantics exclusively from authentic cause triggers.
\section{Conclusion} 
In this paper, we propose \textbf{Causal-EVC}, a cause-grounding captioning framework. To address causal redundancy and spurious correlations, we establish \textbf{EVC-CauseGround}, a pioneering benchmark featuring dense spatio-temporal causal tubes, a curated Causal-Faithfulness Subset, and a convinced metric. 
Technically, Causal-EVC introduces a Motion-guided Causal Localization module with threshold-free heads to decouple foreground causal triggers into a 3D mask. 
Furthermore, we design an Interpretable Sparse Emotion Routing module to deduce class-specific emotion probabilities via sparse prototype routing, enforcing a Counterfactual Contrastive Objective on synthesized background representations to strictly penalize shortcut learning. 
Extensive experiments demonstrate that Causal-EVC achieves the best performance in both text generation and causal faithfulness.

% In this paper, we propose MM-ECPE++, a novel Fine-grained Emotion-Cause Pair Extraction framework for Emotion-Attributed Video Captioning, which performs pairwise extraction of emotional cues and visual causes through multi-round step-by-step refinement. The pairwise extraction promotes each other mutually, enhancing the accuracy and interpretability of emotion perception and helping to decouple the factual and emotional description generation. To be specific, in the 1st round, we propose a concept-aware visual semantic decomposition to refine the visual features through fine-grained concepts mining, including scene, object, and motion concepts. Besides, we propose a visual-guided emotion interpretable learning to filter visual-irrelevant emotional information through visual guiding and VAD-vector constraint. Subsequently, we achieve emotion-cause pair extraction in the 2nd round by cross-coupling the visual and emotional features before and after refinement, and leveraging contrastive loss to achieve semantic forced alignment. Finally, we aggregate visual and emotion-cause semantics into the LLMs-based decoder through a Q-former component for more accurate and vivid emotional captions. Extensive experiments on challenging datasets demonstrate the superiority of our method and each proposed module.

\section*{AI Use Statement}
In this work, we used generative AI tools for polishing the language of the manuscript, writing and debugging portions of the experiment code, and some steps in constructing the dataset. We have not used generative AI tools for the motivation or the production of the reported results and figures. These are carried out directly by the authors. All AI-assisted code and data was tested and verified by running the experiments reported in this paper, and the resulting claims were checked against the run outputs by the authors. We take responsibility for the final content of this work, including the data, text, claims, and artifacts produced with the aid of generative AI.

\bibliography{iclr2026_conference}
\bibliographystyle{iclr2026_conference}

\clearpage

\appendix

\section{Related Work}
\subsection{Emotion-cause Pair Extraction}
Emotion cause analysis has received increasing attention due to its potential to promote emotional exploration. Emotion-cause pair extraction(ECPE) is one of the most representative tasks ~\citep{tang2021graph,li2022ecpec, wang2024enhancing, wang2024enhancing, an2023global,song2025towards,lin2024prompting,10756281}. Xia and Ding ~\citep{xia2019emotion} firstly propose this task to extract emotions and their causes in documents. To this day, recent research has focused more on analyzing emotion-cause pairs in diverse textual scenarios such as articles, stories, microblogs, or conversations. Zhu \emph{et al.} ~\citep{zhu2024knowledge} propose a novel knowledge-guided graph attention network, which designs two guiding principles: inter-clause dependency graph and inter-pair possibility graph, to aggregate features between clause pairs for emotion-clause and cause-clause pair extraction. Chen \emph{et al.} ~\citep{chen2024coarse} propose to utilize the causal discourse knowledge in a knowledge distillation way, which designs a teacher model to learn to predict causal connective words between utterances and then guide a student model in identifying both the fine-grained emotion labels and causal spans. 
Hu \emph{et al.} ~\citep{hu2023emotion} analyze that emotion extraction is more crucial to the ECPE task than cause extraction and propose an approach oriented toward emotion prediction, which aims to fully exploit the potential of emotion prediction to enhance ECPE. Furthermore, they design a synchronization mechanism to share their optimizations in the training process.

Besides, some researchers have extended the ECPE task to the multimodal field ~\citep{li2024multimodal,jin2024improving,lin2024prompting,jin2024improving,liu2024bootstrapping, wang2022multimodal, ma2024extraction, ju2025enhanced,huang2025graph,jin2024improving,jin2024improving2,gao2023vectorized}. 
Wang \emph{et al.} ~\citep{wang2024observe} design a two-stage framework, which first generates emotion-cause aware video captions and then facilitates the generation of emotion causes. With the help of corresponding captions, the model improves the accuracy of pair extractions. However, all existing multimodal ECPE still perform on the text modality while treating image and audio modalities as auxiliary information. In contrast, our work focuses on the cross-modal scenario (visual causes + textual emotions).

\subsection{Emotional Video Captioning}
Emotional video captioning is proposed to solve the problem of boring and soulless captions generated by traditional video captioning model, which first mines emotional clues hidden in videos and collaborates visual contents to generate vivid captions. Wang \emph{et al.} ~\citep{wang2021emotion} firstly propose this task and release a new dataset. They design two independent prediction networks for factual and emotional captioning separately and finally generate emotional descriptions by weighted average scores. Song \emph{et al.} ~\citep{song2022contextual} propose a contextual attention network to recognize and describe the fact and emotion in the video by semantic-rich context learning. Song \emph{et al.} ~\citep{song2023emotion} propose a novel tree-structured emotion learning module to achieve explicit emotion perception. Song \emph{et al.} ~\citep{song2024emotional} incorporate visual context, textual context, and visual-textual relevance into an aggregated multimodal contextual vector to enhance video captioning. Ye \emph{et al.} ~\citep{ye2024dual} propose a dual-path collaborative generation network to dynamically perceive emotions at different time steps and design an emotion-adaptive decoder to solve the problem of overemphasis on the role of emotional guidance. Wang \emph{et al.} ~\citep{wang2025emotion} introduce two learnable prompting strategies: visual emotion and textual emotion prompting, to learn emotional cue representations and further design two levels of objective functions: the ER-sentence level and the AU-word level alignment losses, to facilitate the interaction and alignment. 

All aforementioned works directly perceive the emotional cues from visual contents and ignore that emotional cues have intrinsic motivational causes reflected in the video content. To the best of our knowledge, we are the first work to notice the visual causes of emotions, improving the accuracy and interpretability of emotion perception and effectively alleviating the mutual interference of multiple factors in emotional caption generation.

\section{Theoretical Analysis: From the Information Theory Perspective}
To provide a rigorous theoretical foundation for our Counterfactual Contrastive Objective ($\mathcal{L}_{cf}$), we analyze the fundamental flaws of existing positive-alignment methods (e.g., standard InfoNCE) from an information-theoretic perspective and demonstrate why our interventional approach strictly guarantees causal faithfulness.

\textbf{The Flaw of Standard InfoNCE.} Let $V,Y$ and $C$ denote the random variables for the video, emotion, and true foreground cause, respectively. Traditional EVC models optimize the representation by applying InfoNCE loss between the video representation and the emotion embedding. Mathematically, minimizing the standard InfoNCE loss is equivalent to maximizing the Mutual Information (MI) between the video and the emotion:
\begin{equation}
\min \mathcal{L}_{InfoNCE} \iff \max I(V; Y)
\end{equation}
However, according to our structural causal model, the video is composed of true causes $C$ and the background confounders $U$. By the mutual information chain rule, $I(V; Y)$ can be decomposed as:
\begin{equation}
I(V; Y) = I(C, U; Y) = I(C; Y) + I(U; Y | C)
\end{equation}
In the datasets, the confounder $U$ (e.g., a "party" scene) frequently co-occurs with the emotion $Y$ ("happy") independently of true causes $C$ (e.g., "opening gifts"). Thus, the conditional mutual information $I(U; Y | C)>0$. When a model blindly maximizes $I(V; Y)$, it inevitably maximizes the spurious correlation $I(U; Y | C)$. This explains why existing models suffer from Contextual Leakage. They learn to predict emotions by maximizing the information from the easily accessible background confounders $U$ rather than highly dynamic and harder-to-capture true causes $C$.

\textbf{The Superiority of Counterfactual Intervention.}
To achieve authentic causal reasoning, our goal is to exclusively maximize $I(C; Y)$ while strictly minimizing the spurious leakage $I(U; Y)$. Our framework accomplishes this through the explicitly generated causal mask $\mathcal{M}_c$. We decouple the video into the foreground cause representation $\mathcal{C} = \mathcal{V} \odot \mathcal{M}_c$ and the counterfactual background confounder $U = \mathcal{V} \odot (1-\mathcal{M}_c)$. 

Our proposed Counterfactual Contrastive Objective ($\mathcal{L}_{cf}$) is formulated as:
\begin{equation}
    \mathcal{L}_{cf} = - \mathbb{E} \left[ \log \frac{\exp(\text{sim}(\mathcal{C}, \mathcal{E}) / \tau)}{\exp(\text{sim}(\mathcal{C}, \mathcal{E}) / \tau) + \exp(\text{sim}(U, \mathcal{E}) / \tau)} \right]
\end{equation}
According to the variational bounds of mutual information (e.g., InfoNCE as an MI estimator), optimizing $\mathcal{L}_{cf}$ intrinsically acts on the ratio of conditional densities. By treating the true cause $\mathcal{C}$ as the positive sample and the decoupled confounder $U$ as the strict hard negative sample within the same video context, our objective function tightly bounds the difference in mutual information:
\begin{equation}
    \min \mathcal{L}_{cf} \implies \max \Big( I(\mathcal{C}; \mathcal{E}) - I(U; \mathcal{E}) \Big)
\end{equation}
This formulation yields a profound theoretical guarantee: our model is explicitly penalized if the background confounder $U$ contains any predictive information about the emotion $\mathcal{E}$. To minimize the loss, the optimization process forces the causal mask $\mathcal{M}_c$ to absorb all emotion-relevant semantics into $\mathcal{C}$, driving the spurious mutual information $I(U; \mathcal{E})$ towards zero.

Therefore, compared to traditional positive alignment that inadvertently absorbs environmental biases, our counterfactual intervention rigorously severs the backdoor path $V \rightarrow U \rightarrow Y$. It ensures that the emotion prediction is authentically anchored to the interventional distribution $P(Y | do(C))$, thereby achieving superior causal faithfulness.

% \begin{figure*}[t]        
% \center{\includegraphics[width=1\linewidth] {fig/TPAMI_FIG_2.png}}
% \caption*{Figure A.1: The construction pipeline of the EVC-CauseGround benchmark. (A) Data Source Collection: We collect diverse, in-the-wild videos from EmVidCap-L and MAFW while filtering out overly short clips. (B) Spatiotemporal Visual Cause Annotation: An MLLM-led, tracker-assisted coarse-to-fine pipeline. (C) Human-led Sample Check and Partition: We introduce a Counterfactual Blind Test to divide all samples into causal-redundant and causal-faithfulness subsets.}
% \label{figbench}
% \end{figure*}

\section{More Details of EVC-CauseGround}
\subsection{Comparison with existing benchmark}
In this paper, we construct \textbf{EVC-CauseGround}, a novel benchmark that extends conventional emotion descriptions with a comprehensive suite of fine-grained cause annotations. Table A.1 presents a systematic comparison between existing emotion/causal reasoning datasets and our proposed EVC-CauseGround. Traditional emotion classification and captioning datasets exclusively provide holistic emotion labels or descriptions, fundamentally lacking the annotation of causal triggers, which makes it impossible to assess the true causal reasoning ability. Conversely, while some Emotion-Cause Pair Extraction datasets attempt to address causality, they not only omit comprehensive emotion captions required for generation evaluation, but their causal annotations are also restricted to coarse-grained utterance levels, entirely neglecting fine-grained visual cause localization. Therefore, \textbf{EVC-CauseGround stands as the first benchmark to unify emotion labels, rich emotional captions, dense spatio-temporal visual cause annotations, and a meticulously curated counterfactual subset}. It establishes a solid foundation for comprehensively evaluating the dual capabilities of EVC models: high-quality caption generation and authentic emotional cause reasoning.

\subsection{Details of Data Sources}
The foundation of a robust causal reasoning benchmark relies heavily on the quality and diversity of its video sources. To effectively evaluate the ability to decouple genuine visual emotion triggers from spurious correlations, the selected videos must exhibit highly diverse, open-world scenes (in-the-wild) and sufficient temporal durations. This structural complexity is crucial for providing abundant environmental contexts, acting as potential environmental confounders, against which the model must learn to pinpoint the true causal regions. Driven by these principles, we meticulously curate raw videos from two original datasets.

First, we source a substantial portion of our data from the widely-used EmVidCap~\citep{wang2021emotion} benchmark. The original \textbf{EmVidCap} consists of two subsets: EmVidCap-S and EmVidCap-L. However, we explicitly discard the videos from EmVidCap-S. The videos in EmVidCap-S are exceedingly short (typically under 5 seconds) and feature monotonic scenes. In such cases, the causal action spans nearly the entire video duration, rendering fine-grained spatiotemporal cause annotation virtually meaningless. Therefore, we exclusively select videos from \textbf{EmVidCap-L}. EmVidCap-L comprises videos primarily collected from user-generated content (UGC) platforms, offering a much broader spectrum of daily events and dynamic human interactions. It contains 1523 videos with an average duration of approximately 100 seconds, providing an ideal temporal canvas for fine-grained causal grounding.
Second, to further scale up the benchmark and amplify the diversity, we incorporate a carefully selected subset of videos from \textbf{MAFW}~\citep{liu2022mafw}. MAFW is a large-scale, in-the-wild emotion dataset sourced from various internet platforms (e.g., YouTube and movies) and features an extensive array of unconstrained environments, ranging from extreme sports and natural disasters to bustling streets and crowded parties. The original MAFW dataset contains 10,045 video clips. We filter and select 477 videos from MAFW that exhibit prominent foreground dynamic actions accompanied by complex background scenes.
By amalgamating these two sources, we have successfully collected a total of 2k high-quality emotional videos, which are split into 1500 videos for training and 500 videos for evaluation. Crucially, all selected videos are natively equipped with emotion labels and descriptive emotional captions provided by their original datasets. This inherent semantic richness serves as a reliable ground-truth anchor for our subsequent pipeline to deduce and annotate the precise spatiotemporal causes without suffering from emotion hallucination.

\begin{table*}[t]
\centering
\caption*{Table A.1: \textbf{Comparison of EVC-CauseGround with existing datasets across different domains.} ``V'', ``A'', and ``T'' stand for video, audio, and text modalities, respectively. Our dataset is the first to simultaneously provide fine-grained spatio-temporal causal grounding and a rigorously diagnosed counterfactual subset for real-world emotional videos.}
\label{tab:dataset_comparison}
\resizebox{\textwidth}{!}{
\begin{tabular}{lcccccc}
\toprule
\rowcolor{gray!15}\textbf{Dataset} & \textbf{Modality} & \textbf{Emotion Labels} & \textbf{Emotion Captions} & \textbf{Temporal Grounding} & \textbf{Spatial (Dense Tubes)} & \textbf{Counterfactual Subset} \\
\midrule
\multicolumn{7}{l}{\textit{\textbf{Emotion Classification \& Captioning}}} \\
V-Emo~\citeyearpar{jiang2014predicting} & V, A & \checkmark & $\times$ & $\times$ & $\times$ & $\times$ \\
EmVidCap ~\citeyearpar{wang2021emotion} & V, T & \checkmark & \checkmark & $\times$ & $\times$ & $\times$ \\
\midrule
\multicolumn{7}{l}{\textit{\textbf{Emotion-Cause Pair Extraction (ECPE)}}} \\
ECF ~\citeyearpar{wang2022multimodal} & V, A, T & \checkmark & $\times$ & $\checkmark$ (Utterance-level) & $\times$ & $\times$ \\
MECESD ~\citeyearpar{ma2024extraction} & V, A, T & \checkmark & \checkmark & $\checkmark$ (Utterance-level) & $\times$ & $\times$ \\
\midrule
\multicolumn{7}{l}{\textit{\textbf{Spatio-Temporal Grounding \& Causal Reasoning}}} \\
VidSTG ~\citeyearpar{zhang2020does} & V, T & $\times$ & \checkmark & \checkmark & \checkmark & $\times$ \\
CLEVRER ~\citeyearpar{yi2019clevrer} & V, T & $\times$ & \checkmark & \checkmark & \checkmark & \checkmark (Synthetic) \\
\midrule
\rowcolor{oursburgundy} \textbf{EVC-CauseGround (Ours)} & V, T & \textbf{\checkmark} & \textbf{\checkmark} & \textbf{\checkmark (Frame-level)} & \textbf{\checkmark} & \textbf{\checkmark (Real-world)} \\
\bottomrule
\end{tabular}
}
\end{table*}

\subsection{Spatiotemporal Visual Cause Annotation}
Unlike previous datasets that rely on exhaustive, cost-prohibitive manual drawing, we propose a novel MLLM-led, Tracker-assisted Coarse-to-Fine Pipeline to efficiently generate high-fidelity spatio-temporal causal tubes. This pipeline explicitly circumvents the inherent temporal hallucinations and spatial coordinate drifting typical of current MLLMs when processing long-form videos. The annotation process involves three meticulously designed stages:

\textbf{Temporal Span Grounding via Sequence Grid Prompting.}
Current Video-LLMs struggle with precise temporal localization due to context-length constraints and memory bottlenecks when processing native video streams. To bypass this, we map the temporal dimension into a unified spatial layout. Specifically, we uniformly sample $T=30$ frames from each video, implicitly aligning with the input space of most mainstream EVC models. We then concatenate these frames into a highly structured $6\times 5$ grid image, with each frame explicitly watermarked with a sequence identifier (e.g., $F00$ to $F29$).
We prompt an advanced MLLM (e.g., GPT-5.5~\citep{singh2025openai}) with this grid, alongside the ground-truth emotion label and caption, to reversely deduce the precise temporal window $[t_{start},t_{end}]$ that directly triggers the emotion. Crucially, to prevent the MLLM from exhibiting "lazy behaviors" (i.e., trivially selecting the entire video sequence as the cause), we enforce a strict length-constraining prompt. We mandate that the predicted causal span must not exceed 30\% of the total frames (e.g., max 9 frames). This forces the model to pinpoint the absolute climax or peak moment of the causal action rather than the redundant pre- or post-action context.

\textbf{Spatial Grounding via Peak Motion Anchoring.}
Directly prompting an MLLM to generate bounding boxes across multiple consecutive frames frequently leads to severe spatial hallucinations and bounding box jittering. To ensure pixel-perfect spatial accuracy, we decouple temporal and spatial grounding. Within the localized temporal span $[t_{start},t_{end}]$, we employ the optical flow mechanism to calculate inter-frame pixel displacement. This allows us to automatically identify the peak motion frame, the exact moment where the dynamic causal action is most pronounced.
We then feed only this single, high-resolution peak frame into the MLLM, prompting it to generate a highly confident, static bounding box surrounding the causal agent or action. By reducing the cognitive load to a single image, the MLLM achieves remarkable spatial precision devoid of multi-frame interference.

\textbf{Dense Causal Tube Generation via Zero-shot Tracking.}
A single bounding box is insufficient for dynamic causal reasoning. To extrapolate the spatial annotation across the entire temporal event, we leverage the state-of-the-art zero-shot visual tracker, SAM-3~\citep{carion2026sam}. Utilizing the MLLM-generated bounding box on the peak frame as an initial prompt, SAM-3 bidirectionally tracks the causal object forward to $t_{end}$and backward to $t_{start}$. This seamless integration of MLLM reasoning and traditional visual tracking yields continuous, smooth, and dense spatio-temporal causal tubes, achieving human-level annotation quality at a fraction of the cost.

\section{Experimental Settings}
\textbf{Datasets. }We experiment on two public EVC benchmarks, \emph{e.g.}, EVC-MSVD~\citep{chen2011collecting} and EVC-VE~\citep{jiang2014predicting}. \textbf{EVC-MSVD} contains 240/134 videos and 8,169/4,611 sentences for training/testing by additionally embedding emotion words into the caption sentences of the traditional video captioning dataset MSVD ~\citep{chen2011collecting}. \textbf{EVC-VE} is built based on a traditional emotional video prediction dataset VideoEmotion-8 ~\citep{jiang2014predicting} and is divided into 1,141/382 videos and 19,398/6,527 sentences for training/testing, respectively.

\textbf{Evaluation Metrics. }To evaluate the accuracy of emotion in generated sentences effectively, following previous work~\citep{wang2021emotion}, we consider the emotion word accuracy ${\rm Acc_{sw}}$ ~\citep{wang2021emotion} and emotion sentence accuracy ${\rm Acc_{c}}$ ~\citep{wang2021emotion}. Additionally, to measure the semantic matching degree between the generated sentence and the ground-truth label, we use the standard metrics, \emph{e.g.}, \textbf{BLEU}~\citep{papineni2002bleu}, \textbf{METEOR}~\citep{banerjee2005meteor}, \textbf{ROUGE}~\citep{lin2004rouge}, and \textbf{CIDEr}~\citep{vedantam2015cider}. Moreover, there are two hybrid metrics \textbf{BFS}~\citep{wang2021emotion} and \textbf{CFS}~\citep{wang2021emotion} that combine the emotion evaluation with BLEU and CIDEr metrics, respectively.

\textbf{Implementation Details.} For each video, we sample $T=30$ frames and resize them to $224\times 224$ with central cropping for pixel values. Following ~\citep{wang2021emotion} and ~\citep{song2023emotion}, we build an overall vocabulary of size 32,128 that contains all words in the corpus and set the number of emotion words to $N_w=179$. The embedding dimensions are constructed to $d_E=300$ and $D=1408$. In training, we adopt Qwen-2.5-VL-7B ~\citep{raffel2020exploring} as the language decoder and integrate a LoRA adapter ~\citep{hu2021loralowrankadaptationlarge} with $r=16$ and $\alpha=32$ to align the visual and emotional latent space. We adopt the Adam optimizer ~\citep{kingma2014adam} with a learning rate of 7e-4, and the batch size is set to 4. We set the global motion parameter $\lambda_m$, the penalty parameter $\theta$, the objective function weights $\lambda_e$, $\lambda_{cls}$, $\lambda_{temp}$, $\lambda_{spa}$, and $\lambda_{cf}$ to 0.2, 0.1, 0.8, 0.3, 0.2, 0.2, and 0.1. The maximum length of captions is set to 15 and the size of beam search is set to 4. All experiments are implemented on 4 NVIDIA A800 GPUs.

\section{Baselines} 
We consider several state-of-the-art methods to make comparisons and divide them into two categories: (i) Traditional Captioning Methods and (ii) Emotional Captioning Methods.

\noindent\textit{(i) Traditional Captioning Methods}

1) \textbf{SA-LSTM} (CVPR18') ~\citep{wang2018reconstruction} proposes a reconstruction network to leverage both the forward (video to sentence) and backward (sentence to video) flows. The forward flow produces the sentence description based on the encoded video semantic features and the backward flow reproduces the video features based on the hidden state sequence generated by the decoder.

2) \textbf{CANet} (TMM22') ~\citep{song2022contextual} proposes a contextual attention network to recognize and describe the fact and emotion by semantic-rich context learning, which first extracts visual and textual features from both video and previously generated words and then applies the attention mechanism to capture informative contexts for captioning.

3) \textbf{VideoBLIP} (EMNLP24') ~\citep{yu2024eliciting} proposes a training paradigm that induces in-context learning over video and text by capturing key properties of pre-training data found by prior work to be crucial for improving the ability of LLMs to generate video descriptions.

\smallskip
\noindent\textit{(ii) Emotional Captioning Methods}

1) \textbf{FT} (TMM21') ~\citep{wang2021emotion} introduces a visual emotion analyzer to detect implicit emotional cues and a dual-stream network that integrates emotional and factual semantics for caption generation by a weighted sum operation.

2) \textbf{EPAN} (MM23') ~\citep{song2023emotion} introduces a tree-structured emotion repository to enable hierarchical emotion perception and an emotion-prior awareness network to achieve the explicit and fine-grained emotion perception by an emotion masking mechanism and then decode the emotional caption by exploiting the multimodal semantic cues.

3) \textbf{VEIN} (TIP24') ~\citep{song2024emotional} designs a vision-based emotion interpretation network, which first models the emotion distribution over an open psychological vocabulary and then incorporates visual context, textual context, and visual-textual relevance into an aggregated multimodal contextual vector to enhance video captioning.

4) \textbf{DCGN} (MM24') ~\citep{ye2024dual} propose a framework to leverage fact-emotion collaborative learning to address the challenge of dynamic emotion changes during caption generation. 

5) \textbf{MM-ECPE} (MM25')~\citep{ye2025multi} focus on the importance of emotional
causes in emotional exploration and propose a multi-round mutual learning network to jointly extract emotion-cause pairs for LLM-based caption generation.

6) \textbf{HEART} (TAFFC26')~\citep{han2026heart} introduces a hierarchical semantic extraction module to decompose visual content into entity-, action-, and event-level representations, and a temporal pyramid module to capture temporal dependencies for temporally coherent captioning.

7) \textbf{GLM-EER} (ESWA26')~\citep{ma2026glm} designs a global-local memory module to learn abstract emotional representation across time dimensions and a Emotion Evaluation Refinement module to enhance emotion learning by dynamically adjusting cross-entropy loss weights.

8) \textbf{SAGML} (Arxiv26')~\citep{wang2026adaptive} proposes an adaptive EVC framework via affective heterogeneous graph and multi-task language modeling, which constructs a soft affective heterogeneous graph to allow visually supported lexical emotions to remain recoverable.

\section{Human Evaluation}
Due to the highly subjective nature of the EVC task, it is a crucial criterion for evaluation whether the generated descriptions conform to human preferences. Therefore, we add human evaluation to fully measure the quality of emotional descriptions. Inspired by the work ~\citep{wang2021emotion}, we designs four metrics, including (1) emotion accuracy: assess the accuracy of emotional expression, (2) relevance: assess whether the generated description is relevant to the video content, (3) coherence: assess the coherence and readability of the description, and (4) usability: how useful would the description be for a person (especially a blind person) to understand what is happening in the video. The score of each metric ranges from 1 to 10. Subsequently, we invite 20 participants with excellent English skills to score a subset of 50 video-caption pairs on these four metrics and calculate the average score on each metric. We make a comparison with two types of models on human evaluation: the state-of-the-art models on EVC and several large vision-language models (LVLMs).

\noindent\textbf{Comparison with SOTA methods. }As shown in Table A.2, compared with the SOTA methods for EVC, our model is significantly ahead in all four metrics. Our proposed MCSL captures fine-grained cause localization, thereby generating more accurate and diverse vocabulary. Furthermore, our proposed ISER utilizes VAD constraints to enhance the interpretability of emotions, making the generated descriptions more easily empathized with and understood by humans. These show that our model does not simply imitate the reference sentences, but learns the representation of emotional and factual semantics.

\noindent\textbf{Comparison with LVLMs. }As shown in Table A.2, despite the outstanding performance of LVLMs, our model still outperforms them in human evaluations. Due to their powerful ability in objective understanding and generation, LVLMs could produce a wide variety of descriptions. However, limited by insufficient emotion understanding and the influence of visual hallucinations, they sometimes generate factual and emotional contents irrelevant to the video, leading to lower human ratings. Conversely, our model filters out emotion-irrelevant visual information and visual-irrelevant emotional information through fine-grained visual and emotional refinement, thereby generating emotional descriptions that are more closely aligned with the video content.

\begin{table}[h]
\centering
\caption*{Table A.2: {The comparison with SOTA methods on human evaluations.}}
\resizebox{0.95\columnwidth}{!}
{
\begin{tabular}{c|cccc}
\toprule
{Metric} & Accuracy&Relevance&Coherence&Usability\\
\midrule
\rowcolor{gray!15}VEIN~\citeyearpar{song2024emotional}& 5.47 & 5.18 & 6.27 & 5.04 \\
EPAN~\citeyearpar{song2023emotion}& 6.24 & 5.97 & 7.13 & 5.76\\
DCGN~\citeyearpar{ye2024dual}& 6.88 & 6.43 & 7.68 & 6.32\\
VideoBLIP~\citeyearpar{yu2024eliciting}& 5.91& 5.54 & 7.20 & 4.56\\
MM-ECPE~\citeyearpar{ye2025multi}& 6.40 & 6.75 & 8.04 &7.36 \\
\midrule
GPT-4o~\citeyearpar{hurst2024gpt}  & 6.98  &6.04  & 7.24  &6.62     \\
VideoLLaMA3~\citeyearpar{zhang2025videollama}  &7.22   & 6.20 &7.72   & 6.84     \\
\rowcolor{oursburgundy}{\textbf{CausalEVC(Ours)}}& \textbf{7.52} & \textbf{6.85} & \textbf{8.16} & \textbf{7.58}\\
\bottomrule
\end{tabular}}
\end{table}

\end{document}